\documentclass[preprint,review,12pt]{elsarticle}

\usepackage{booktabs}
\usepackage{amssymb}
\usepackage{amsmath}
\usepackage{pifont} 
\newcommand{\xmark}{\ding{55}}
\newcommand{\cmark}{\ding{51}}
\usepackage{xcolor}
\usepackage{multirow}
\usepackage{makecell}

\newcommand{\eg}{\emph{e.g.}}

\journal{Nuclear Physics B}

\begin{document}

\begin{frontmatter}



\title{Take a Prior from Other Tasks for Severe Blur Removal}

\author[label1,label4]{Pei Wang}
\author[label1,label3]{Danna Xue}
\author[label1]{Yu Zhu}
\author[label2]{Jinqiu Sun}
\author[label1]{Qingsen Yan}
\author[label4]{Sung-eui Yoon}
\author[label1]{Yanning Zhang}

\address[label1]{{School of Computer Science and Engineering, Northwestern Polytechnical University}, 
            {Xi'an},
            {China}}
\address[label2]{{School of Astronautics, Northwestern Polytechnical University},
            {Xi'an},
            {China}}
\address[label3]{{Computer Vision Center, Universitat Autònoma de Barcelona},
            {Barcelona},
            {Spain}}
\address[label4]{{School of Computing, Korea Advanced Institute of Science and Technology},
            {Daejeon},
            {Korea}}
\begin{abstract}
Recovering clear structures from severely blurry inputs is a challenging problem due to the large movements between the camera and the scene. Although some works apply segmentation maps on human face images for deblurring, they cannot handle natural scenes because objects and degradation are more complex, and inaccurate segmentation maps lead to a loss of details. For general scene deblurring, the feature space of the blurry image and corresponding sharp image under the high-level vision task is closer, which inspires us to rely on other tasks (e.g. classification) to learn a comprehensive prior in severe blur removal cases. We propose a cross-level feature learning strategy based on knowledge distillation to learn the priors, which include global contexts and sharp local structures for recovering potential details. In addition, we propose a semantic prior embedding layer with multi-level aggregation and semantic attention transformation to integrate the priors effectively. We introduce the proposed priors to various models, including the UNet and other mainstream deblurring baselines, leading to better performance on severe blur removal. Extensive experiments on natural image deblurring benchmarks and real-world images, such as GoPro and RealBlur datasets, demonstrate our method's effectiveness and generalization ability.
\end{abstract}






\end{frontmatter}


\section{Introduction}
\label{introduction}
Image blur is caused by relative movements between the imaging device and objects, including camera shaking and object movements. To eliminate the blur, the image deblurring task has been studied actively, yet is a classical ill-posed problem in computer vision. Given a blurry image $\mathbf{y}$, the deblurring process can be formulated as:
\begin{equation}
\label{eq1}
\mathbf{y} = \mathbf{k} \otimes \mathbf{x} + \mathbf{n},
\end{equation}
where $\otimes$ represents the convolution operator, $\mathbf{k}$ is blur kernel, $\mathbf{x}$ denotes the sharp image, and $\mathbf{n}$ is assumed as the additive white Gaussian noise.

Previous works develop various techniques to improve deblurring performance \cite{pan2016blind,pan2014deblurring,zhang2023learning}. They effectively restore a clear image in some simple conditions, \textit{e.g.}, uniform blur, or specific scenarios (human face or text image). However, it is a challenge to handle severe blur, which indicates that the blur kernels in Eq. (\ref{eq1}) have very large sizes or complex shapes. Regions with severe blur not only lose considerable details but also have ambiguous semantics. The contexts of these areas are confusing, making it difficult to distinguish each object in terms of category and structure. Current networks capture features for non-uniform deblurring by multi-scale \cite{tao2018scale-recurrent,nah2017deep,liu2021multi,WANG2021108082,peng2020joint}, multi-patch \cite{zhang2019deep,suin2020spatially}, or multi-stage \cite{Zamir2021MPRNet,chen2021hinet,Kim2022MSSNet} architectures, which mainly aims to enlarge the receptive fields. However, the learning behavior of the networks in these severe blur regions is difficult to predict due to the ambiguous semantics. Furthermore, these methods rely on the information encoded by the blurry inputs, which is insufficient to recover rich details in severe blur situations. For scenes with real complex blur, such as fast motion-induced blur, deblurring based only on blurred images is extremely hard.

In order to solve the problem of semantic confusion caused by severe blur, some works \cite{Shen2018DeepSF,Shen2020ExploitingSF,Yasarla2020DeblurringFI} have attempted to use semantics for human-face deblur by directly regarding the segmentation maps as the input \cite{Shen2020ExploitingSF} or the supervision \cite{Yasarla2020DeblurringFI} of the trainable models. Because of the face's fixed component and simple structure, they have achieved good performance in this specific scenario. However, the segmentation map only contains global categories and neglects the detailed structure, which cannot handle complex regions, such as eyes and lashes. Thus, it's hard to apply them to general natural scenes that include complicated components and severe blur. In this paper, we propose to learn a semantic prior to assisting the deblurring process. There are three questions beforehand: \textbf{\textit{(1) what kind of priors can benefit image deblurring? 
(2) how to learn the favorable priors? 
(3) how to effectively employ the priors?}} For issue (1), the proper semantic prior has a strong capacity for severe blur removal. Different from the previous works \cite{Shen2018DeepSF,Shen2020ExploitingSF} that rely on the segmentation map to only provide global contexts, we deem a prior to incorporate both global and local structures as a bridge to recover a clear image. The prior with semantics can assist the framework to discriminate the different semantic regions. Because the local structures, such as edges and textures, are significant in the image deblurring task, the priors also include as many as possible clear local details simultaneously.
For issue (2), considering that high-level vision tasks, such as classification or segmentation, can naturally offer rich semantics and identify the specific object. We derive a learning strategy by leveraging high-level vision tasks to learn semantics and adopting the ground truth image to provide sharp local information. 
For issue (3), we integrate the priors into existing deblurring frameworks by a semantic attention-based module, which maximally utilizes the semantic prior and further improves the performance of the current methods.

In this paper, we investigate semantics from high-level vision tasks, e.g., classification, to guide severe blur removal. We construct a learning strategy based on knowledge distillation to produce a reasonable semantic prior. 
\textbf{Combining the privilege distillation and the high-level vision tasks enables the priors to mimic the capacity to discriminate different objects in the severe blur condition. The priors are supervised by sharp images, yielding clear global and local semantic knowledge, which improves deblurring ability.}
In addition, we propose a prior embedding layer with feature aggregation and transformation to utilize the learned priors effectively. Thanks to the priors with rich global semantics and local sharp details, the severe blur can be significantly removed and potential details can be recovered. Experiments on benchmarks demonstrate the effectiveness and the generalization ability of our method, which improves the performance of the UNet as well as other mainstream deblurring methods, shown in Section \ref{sec:experiments}. 
Our work takes meaningful steps toward making deblurring networks semantically aware and provides a unique representation with interpretable behavior. The contributions are summarized as: 
\begin{itemize}
\item To the best of our knowledge, our method is the first severe blur removal framework in the general scene with learning a comprehensive sharp semantic prior guidance. 
\item We exploit a semantic prior learning strategy under the knowledge distillation by the cross-level feature transferring constraints. 
\item A semantic prior embedding layer with feature aggregation and attention transformation are proposed to utilize the semantic prior to the image deblurring framework effectively.
\item Our approach has a robust generalization ability to boost the performance of the current deblurring methods, including UNet \cite{depthLi}, DeepDeblur \cite{nah2017deep}, MPRNet \cite{Zamir2021MPRNet}, and MIMO-UNet \cite{cho2021rethinking}.
\end{itemize}

The remaining parts of the paper are structured as follows. Related image deblurring algorithms are summarized in Section \ref{sec:related}. We elaborate on the framework in Section \ref{sec:approach}. Section \ref{sec:experiments} describes the details of the experimental setting and the results. The several ablation studies and some discussions are in Section \ref{sec:ablation}. Finally, we conclude our work in Section \ref{sec:conclusion}.

\section{Related Works}
\label{sec:related}
\subsection{General image deblurring}
The traditional methods for deblurring designed several priors of the natural sharp images to reduce the resolution space, such as $L_0$ \cite{xu2013unnatural} and dark channel \cite{pan2016blind}. As most priors are hand-crafted and based on limited observations of specific scenes, these approaches do not generalize well to handle various blurry images. 
With the development of the deep-learning, many works \cite{tao2018scale-recurrent,nah2017deep,zhang2019deep,suin2020spatially,chen2021hinet,sun2015learning,gong2017motion} try to use the learning approach to replace the optimization process. 
\cite{sun2015learning,gong2017motion} replace the kernel iteration with CNN blocks and input the predicted kernel flow and blurry image to another network 
to restore the latent image. It acquires great progress compared to the optimization methods. Subsequently, the end-to-end learning framework has been widely used, which directly learns a mapping from the blurry image to the sharp image. Multi-scale framework \cite{nah2017deep} greatly improves the deblurring quality since blurry images in different resolutions can provide features with various characteristics. \cite{tao2018scale-recurrent} and \cite{gao2019dynamic} improve the multi-scale framework proposed by Tao and Gao \textit{et al.} with parameter sharing. \cite{zhang2019deep} and \cite{suin2020spatially} use multi-patch to reduce the information loss during the up/down sampling stages.
However, these methods mainly focus on network design. Although they developed various useful structures, they only consider the information encoded by the input blurry images. This may not be enough for severe blurry images with significant information loss. To alleviate the severe blur, we propose a framework that uses the information not only from the blurry image but also the other tasks, such as classification.

\subsection{Deblurring with additional information} 
Several approaches introduce additional information to improve the challenging scenes in image deblurring. 
\cite{jiang2020learning} and \cite{Shang_2021_ICCV} leverage the event data in single and video image deblurring tasks, providing a new perspective on deblurring. 
\cite{Shen2018DeepSF,Shen2020ExploitingSF,Yasarla2020DeblurringFI} generate segmentation maps by a face parsing module and regard the maps as the global contexts of the input to deblurring. These methods perform well on human-face images with relatively fixed patterns. However, natural scene images have much more diverse components, and the locations and shapes of the moving objects are changing from image to image.
\cite{depthLi} proposes a depth-guided image deblurring model, which generates an initial depth map, then refines the depth and deblurring in a joint learning way. Similar to \cite{Shen2018DeepSF, Shen2020ExploitingSF}, the depth map also can be seen as a global prior, which only contains the shape and spatial information.
However, image details are critical for deblurring tasks, and recovering them from blurry images is a huge challenge. Thus, unlike the segmentation or depth guidance, we investigate a sharp semantic prior to severe blur removal that combines the global context and local structure information. 

\subsection{Knowledge distillation}
Knowledge Distillation (KD) refers to the training process that learning a smaller student network under the supervision of a larger teacher network. FitNet \cite{romero2014fitnets} is proposed to train the student models with the feature map of teacher models instead of the \textit{logit}.
\cite{zagoruyko2016paying} applies feature distillation on the attention map of networks at different layers. 
Except for high-level vision tasks some image restoration methods \cite{liu2020residual} also use KD to get a small yet efficient model for mobile device applications. Most existing methods provide students and teachers with the same input since they aim to obtain a student with the same abilities as teachers. For supervised image restoration tasks, there are pairs of degradation and ground truth images for training. What if directly distilling the valuable knowledge from the GT image to help the degraded image recovery? \cite{NIU202169} shares one network for distillation with sharp and blurry image inputs. It is unclear what kind of information is generated by the network since it's difficult to learn two kinds of distribution (sharp and blur) simultaneously with one model. \cite{lee2020learning} employs distillation to extract the privileged information from HR images to guide the LR image's super-resolution process. It gives us an example of distilling high-quality information from the GT to assist in image recovery. Following \cite{lee2020learning}, we propose a deblurring framework with two separate networks, aiming to learn semantic priors for transferring knowledge from the sharp image.

\section{Approach}
\label{sec:approach}
In this paper, we propose a semantic prior-guided image deblurring framework to learn and leverage a sharp semantic prior for recovering potential details from severely blurry images. The overview of the method is shown in Figure \ref{fig:overview}. Assisted by high-level tasks, such as classification or segmentation, we design a Semantic Prior Learning branch (SPL) to produce semantic priors with both global contexts and sharp local structures from the input blurry image. Then, the learned prior is fed into the Image Deblurring branch (IDe) through a Semantic Prior Embedding layer (SPE) to guide the blur removal. We use an encoder-decoder structure to build the deblurring branch for convenience. Moreover, the following experiments in Section \ref{resluts} demonstrate that replacing the encoder-decoder with other state-of-the-art deblurring networks \cite{nah2017deep,Zamir2021MPRNet,cho2021rethinking} also performs well.

\subsection{\textbf{Learning semantic priors based on knowledge distillation} }
\label{sec:KD}
\begin{figure*}
    \centering
    \includegraphics[width=5.4in]{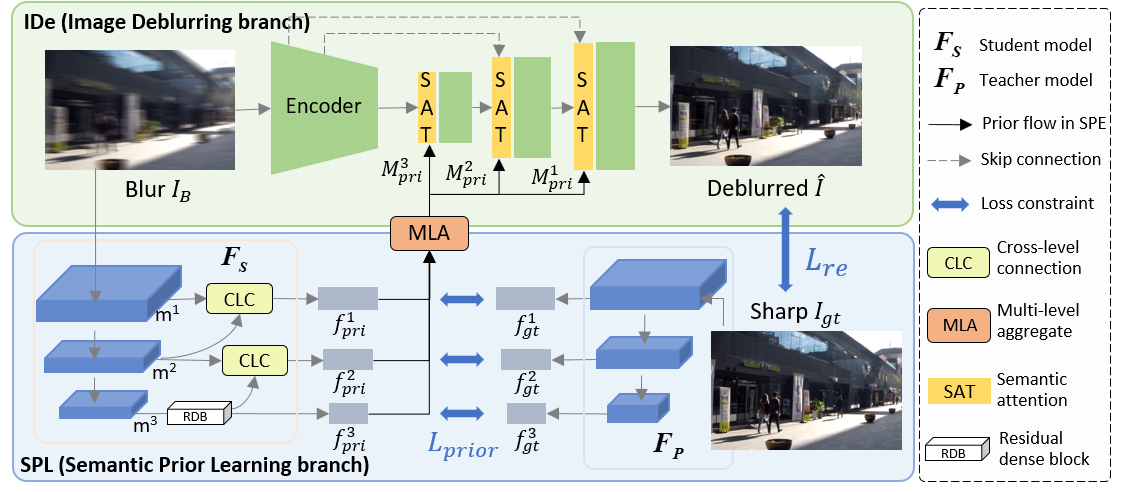}
    \caption{Semantic prior-guided image deblurring consists of an Image Deblurring branch (IDe) and a Semantic Prior Learning branch (SPL). The Semantic Prior Embedding layer (SPE) pushes the priors into IDe, which contains a Multi-Level Aggregate (MLA) and a Semantic Attention Transformation module (SAT). $\mathcal{L}_{re}$ indicates the reconstruction loss between sharp and deblurred images. $\mathcal{L}_{prior}$ represents the semantic prior learning loss.}
    \label{fig:overview}
\end{figure*}

As we mentioned in Section \ref{introduction}, we need to learn a prior that has global contexts and sharp local structures to help the severe blur removal. We build a semantic prior learning branch (SPL) to capture sharp semantic priors from the high-level vision tasks (\eg, classification, segmentation). 
As shown in Figure \ref{fig:overview}, we feed a blurry image $I_B$ into a prior extraction module $F_S$ to generate prior $f_{pri}$, which is supervised by a high-level vision model $F_P$ with a sharp image input $I_{gt}$. The outputs of $F_P$, noted by $f_{gt}$, contain the abundant referenced semantic information. Therefore, our prior $f_{pri}$ can acquire explicit global semantic knowledge, improving deblurring ability. 
\subsubsection{\textbf{Prior types}}
Generally speaking, our framework supports multiple prior types $f_{pri}$ learning from various vision tasks through the SPL branch, such as images and features. 
If the prior is images, such as the segmentation map, we use a semantic segmentation model to build $F_P$. In this condition, $f_{pri}$ is similar to the guidance in \cite{Shen2018DeepSF} and \cite{Shen2020ExploitingSF}, which can be grouped by \textit{image-level prior}. However, the restored images with this image-level prior still contain obvious residual blur (See Figure \ref{fig:seg_prob}), which reveals that simply using segmentation maps is not an effective approach for blurry image recovery. Because they only provide global information about objects and ignore the details that are very important for image restoration tasks, such as the textures.  
As introduced in Section \ref{introduction}, our priors need to include as much sharp local detail as possible to reconstruct the latent image. Thus, image-level priors are not very suitable for our task.

To solve the above issue, we learn a \textit{feature-level prior}, whose supervision $f_{gt}$ comes from the intermediate layers of the $F_P$. The feature-level prior can provide more abstract context to assist object recognition, as well as rich structures to complement image details through the multi-scale learning model $F_S$.  
As shown in Fig. \ref{fig:overview}, the proposed SPL branch has three scales. For convenience, according to the multi-scale structure, the semantic prior $f_{pri}$ and supervised signal $f_{gt}$ also can be denoted by $f_{pri}^i$ and $f_{gt}^{i}$, respectively, where $i$ denotes the level of the features.

\subsubsection{\textbf{Semantic Prior Learning Strategy}}
The semantic prior $f_{pri}$ is produced from the blurry image through the prior extraction module $F_S$. We propose a distillation-based semantic prior learning strategy to align the feature distribution between $f_{pri}$ and $f_{gt}$, accompanied by a \textbf{cross-level feature distillation loss $\mathcal {L}_{prior}$} to learn the abundant semantic information and sharp structures. The high-level vision model $F_P$, \textit{i.e.} the teacher model, captures the precise structure and semantic information from a sharp image. The student model $F_S$ aims to learn the robust and adequate privileged knowledge from the input blurry images under the supervision of the sharp image features generated by $F_P$. The detailed structure of the Semantic Prior Learning branch is shown in the bottom part of Figure \ref{fig:overview}.

To learn the local sharp structures, we employ an enhanced L2 norm distance in feature distillation, named Hierarchical Context Loss (HCL) \cite{chen2021distilling}, which minimizes the mean-square-error in multi-scale space between the semantic prior and supervised signal. Compared with the ordinary L2 distance, the context loss considers the feature distance at different resolutions, which maintains the hierarchical scale information in the local feature space. The formula for feature distillation is as follows:
\begin{equation} 
\label{eq:fea}
   \begin{aligned} 
   \mathcal{L}_{prior\_S} &= \frac{1}{L} \sum_{i=1}^{L}\frac{1}{W^iH^i}{\mathcal {L}_{HCL}(m^{i},f_{gt}^{i})} \\
   &= \frac{1}{L} \sum_{i=1}^{L} \frac{1}{W^iH^i} \sum_{k}{\left\| avg^k(m^{i})-avg^k(f_{gt}^{i}) \right\|_2},
   \end{aligned}
\end{equation}
where $m^{i}=F_{S}(I_B)_i$ represents the output features from the prior extraction module, and $f_{gt}^{i}=F_{P}(I_{gt})_i$. $I_B$ and $I_{gt}$ are the blurry input and ground truth. $i$ denotes the \textit{$i_{th}$} level of the features or structures. $L=3$ means the total number of the scale. $\left\{ {W^i,H^i}\right\}$ represent the width and height of the features, respectively. $avg^k$ represents the average pooling operator with the factor $k$, and $k$ belongs to \{1,2,4\}, which represents calculating the distance in the original, 1/2, and 1/4 scale. 

\begin{figure}
    \centering
    \includegraphics[width=4.5in]{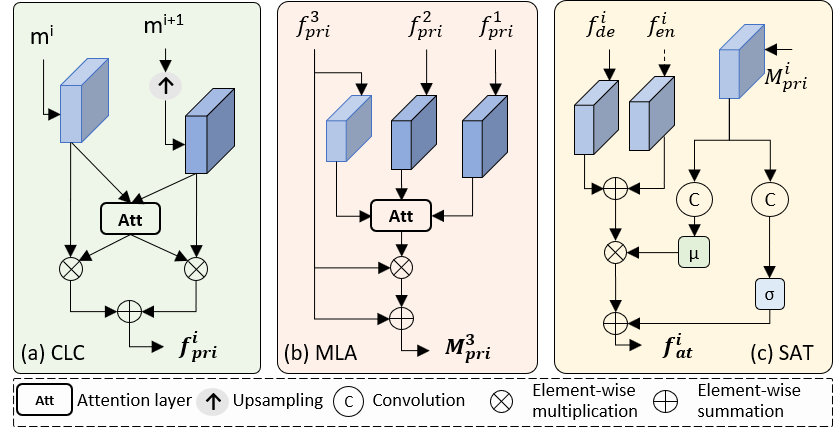}
    \caption{The detail structures of the Cross-Level Connection (CLC), Multi-Level Aggregate (MLA), and Semantic Attention Transformation (SAT). }
    \label{fig:MLA_FT}
\end{figure}

However, \textit{on one hand}, employing the loss in Eq. (\ref{eq:fea}) only distills the features within the same levels, such as $m^{1}$ and $f_{gt}^{1}$. \textit{On the other hand}, the features in the shallow-level cannot cover the deep-level semantic information, and the priors learned from this strategy are insufficient in abstract semantics representations.
Recently, Chen \textit{et al.} \cite{chen2021distilling} discover that the student's deep-level stage has the great capacity to learn useful information from the teacher’s shallow-level features.
Inspired by this, we propose a Cross-Level Connection module (CLC) to pull the abstract features from the deep-level ($m^{i+1}$) to the shallow-level ($m^i$), 
which improves the $F_P$'s constraint ability and enhances the semantics of the shallower features in $F_S$. 
The detailed structure of CLC is shown in Figure \ref{fig:MLA_FT}(a). We employ an attention layer to construct the CLC:
\begin{equation}
\label{eq:clc}
   \begin{aligned}
    f_{pri}^{i}  &= f_{att}^i \otimes m^i + f_{att}^i \otimes (m^{i+1}_{\uparrow}),
   \end{aligned}
\end{equation}
where $i\in (0,1)$, and $f_{pri}^{3} = m^3$. $f_{att}^i= conv(cat(m^i,m^{i+1}_\uparrow))$ indicates the attention map, which can be regarded as a re-weighting parameter to modulate the current feature $m^i$. To maintain the original dimension, we adopt an upsampling layer ($\uparrow$) to transfer the shape of $m^{i+1}$ to $m^i$. $cat$ and $conv$ represent the concatenate and convolution operations. $\otimes$ indicates the element-wise multiplication. Therefore, the enhanced cross-level feature distillation loss $\mathcal {L}_{prior}$ is shown:
\begin{equation}
\label{eq:krf}
    \mathcal{L}_{prior} = \frac{1}{L} \sum_{i=1}^{L}\frac{1}{W^iH^i} \mathcal {L}_{HCL}(f_{pri}^{i},f_{gt}^i).
\end{equation}

With Eq. (\ref{eq:clc}) and (\ref{eq:krf}), the shallow-level features in the teacher network constrain the deeper features of the student. For instance, $m^2$ from the student $F_S$ is connected with $m^3$ to imitate the $f_{gt}^2$ from teacher $F_P$. Therefore, $m^3$ learns the residual information between the $m^2$ and $f_{gt}^2$. \textbf{With the cross-level feature distillation, the semantic prior can obtain not only the rich sharp details from the shallow level features $m^i$, but also the abstract semantics from the deeper scales $m^{i+1}$.}
\subsection{\textbf{Image deblurring branch and semantic prior embedding}}

Through the semantic learning branch (SPL), we obtain a comprehensive semantic prior, which contains abundant semantic information and sharp structures of the latent image. To leverage the valuable knowledge of the prior in the Image Deblurring branch (IDe), we propose a Semantic Prior Embedding layer (SPE) for each level of the IDe's decoder, which is shown in the middle of Figure \ref{fig:overview}. Here, we use the simplest UNet as the backbone of the IDe branch to verify the effectiveness of the proposed semantic prior. It can be replaced by multiple other deblurring models, such as DeepDeblur \cite{nah2017deep}, MPRNet \cite{Zamir2021MPRNet}, and MIMOUNet\cite{cho2021rethinking}. 

The SPE consists of a Multi-Level Aggregate module (MLA) and a Semantic Attention Transformation module (SAT). 
To maximize the ability of the multiple scales priors for deblurring, we propose an MLA module in SPE containing an attention layer. MLA and CLC have similar structures but for different use.
As shown in Figure \ref{fig:MLA_FT}(b), MLA aggregates the learned multi-scale priors with different dimensions to obtain a semantics merged prior $M_{pri}^i$, which can be embedded to the deblurring branch: 
\begin{equation}
    \begin{aligned} 
    M_{pri}^i &= \underset{t\in \{1,...,L\},t\neq i}{MLA}\left (  f_{pri}^i,{f}_{pri}^t \right ) \\
              &= conv (cat(f_{pri}^i,{f}_{pri}^t)) \otimes f_{pri}^i + f_{pri}^i,
    \end{aligned}
\end{equation}
where ${f}_{pri}^t ({t\in \{1,...,L\},t\neq i}$) represents the features of the remaining levels in addition to the $i_{th}$ level. $\otimes$ indicates the element-wise multiplication. we employ linear interpolation to make the resolution consistent. 

As shown in Figure \ref{fig:MLA_FT}(c), the SAT is proposed to integrate the aggregated semantic priors $M_{pri}^i$ to the IDe branch along the features level and guide the deblurring process. Through the proposed semantic attention module, the features extracted by the deblurring network are assigned different weights to image regions with different semantic regions and blurring degrees, so as to better deal with global inconsistent blurring. 
The two additional convolution blocks are used to calculate the transformed scale $\mu=conv1(M_{pri}^i)$ and shift $\sigma=conv2(M_{pri}^i)$. With a skip connection operation, the deblurred features of the encoder $f_{en}^{i}$ and decoder $f_{de}^{i}$ are modulated by the semantic priors $M_{pri}^i$, and the modulated feature $f_{sa}^i$ with effective semantic information can be generated by:
\begin{equation}
    f_{sa}^i = \mu \otimes (f_{de}^{i}+f_{en}^{i}) + \sigma ,
\end{equation}
where $f_{de}^{i}$ and $f_{en}^{i}$ represent the features of decoder and encoder of the $i_{th}$ level. Then, $f_{sa}^i$ is fed into the following stages to recover the latent image $\hat{I}$. 

\subsection{\textbf{Training total loss}} 
In addition to the semantic prior loss $\mathcal{L}_{prior}$, we employ another image reconstruction loss $\mathcal{L}_{re}$ to train the whole deblurring framework, which consists of a $\mathcal{L}_{1}$ distance loss and a perceptual loss $\mathcal{L}_{vgg}$. $\mathcal{L}_{1}$ focus on the pixel-level reconstruction and $\mathcal{L}_{vgg}$ pays more attention to the human perception view. Therefore, the training total loss can be formulated by:
\begin{equation}
   \begin{aligned}
  \mathcal{L}_{total} &= \underbrace{\mathcal{L}_{1} + \alpha\mathcal{L}_{vgg}}_{\mathcal{L}_{re}}+\beta\mathcal{L}_{prior},
   \end{aligned}
\end{equation}
where $\alpha=0.01$ and $\beta=0.1$ are the hyper-parameters to balance the three losses. Specifically, the detailed formulation of $\mathcal{L}_{1}$ and $\mathcal{L}_{vgg}$ are as follows:
\begin{equation}
  \begin{aligned}
\mathcal{L}_{1} &=\frac{1}{HW}\sum_{h=1}^{H}\sum_{w=1}^{W}\left\|I_{{gt}(h,w)}-\hat{I}_{(h,w)} \right\|_1, \\
\mathcal{L}_{vgg} &= \frac{1}{H_pW_pC}\left\|\sum_{h=1}^{H_p}\sum_{w=1}^{W_p}\sum_{c=1}^{C} {\phi_{(h,w,c)}(I_{gt})-\phi_{(h,w,c)}(\hat{I})} \right\|_2,\\
  \end{aligned}
\end{equation}
where $I_{gt}$ and $\hat{I}$ represent the ground truth and deblurred image, respectively. $\phi$ denotes the operation of extracting the feature maps from the intermediate layer of the VGG model \cite{simonyan2014very}, which has been well-trained on the ImageNet dataset \cite{deng2009imagenet}. Here, we use activations from $VGG_{3,3}$ convolutional layer as $\phi$. $C$ denotes the channel numbers of the $\phi$. $\{H,W\}$ and $\{H_p,W_p\}$ indicate the height and width of the images and features maps.

\section{Experiments}
\label{sec:experiments}
\subsection{Implementation details}
In recent years, U-shape architecture becomes the major structure for image deblurring tasks, such as \cite{Zamir2021MPRNet,Kim2022MSSNet,depthLi,cho2021rethinking}, since the features learned by the U-shape network will be more beneficial for deblurring by producing a large receptive field and the proper resolution. Therefore, we mainly conduct experiments on the deblurring models with U-shape architectures to demonstrate the idea. To verify the generalization of the proposed priors, we also provide the results on the other deblurring baselines, such as multiple ResBlocks stacked model \cite{nah2017deep}. 
The $F_S$, $F_P$, and the IDe branch have the same number of layers $L=3$. In the following experiments, we employ the models of image classification \cite{ridnik2021asymmetric,ijcai/context} and segmentation \cite{zhu2019improving} as the high-level vision task models $F_P$, which have been well-trained on the sharp images of OpenImages \cite{OpenImages2} and Cityscapes \cite{cordts2016cityscapes} datasets, respectively.

All experiments are implemented in PyTorch and evaluated on a single NVIDIA Geforce RTX 3090 GPU. To train our network, we randomly crop input images to $256\times256$ pixel size. The batch size is set to 10 during training. The AdamW solver ($\beta_1=0.9$, $\beta_2=0.999$) is used to train models with the learning rate of $2\times10^{-4}$, which is steadily decreased to $1\times10^{-6}$ using the cosine annealing strategy. Since our key idea is introducing the semantic prior to assist the deblurring process, we train the models with and without the prior for 700 epochs to verify the effectiveness.

\subsection{Datasets and evaluation metrics}
We train and evaluate our models on two widely-used synthetical datasets, GoPro \cite{nah2017deep} and HIDE \cite{Shen2019HumanAwareMD}.  
We also evaluate our models on a real-world dataset, RealBlur \cite{rim2020real}, whose blurry images are captured under real-world blurry conditions.
\textbf{GoPro} dataset \cite{nah2017deep} consists of 2,103 training and 1,111 testing image pairs with various dynamic motion blur. The images are generated by recording video clips with high shutter speeds, then averaging consecutive 7-13 frames to simulate blur caused by a slow shutter. 
\textbf{HIDE} \cite{Shen2019HumanAwareMD} is specifically collected for human-aware motion deblurring. The training and testing sets contain 6,397 and 2,025 images. The blur in the HIDE dataset is more severe due to the aggregating of 11 consecutive frames, leading to huge movements in every blurry image. \textbf{RealBlur} dataset \cite{rim2020real} has two subsets: (1) RealBlur-J is formed with the camera JPEG outputs, and (2) RealBlur-R is generated offline by applying white balance, demosaicking, and denoising operations to the RAW images. We use RealBlur-J which has 3,758 training and 980 testing images. We train and test our model on RealBlur-J to verify the ability of our model to process real-world blurry images in this paper.

We evaluate the quality of deblurred images with common pixel-level distortion metrics: PSNR and SSIM, and a perceptual quality metric: LPIPS.

\subsection{Comparisons with previous methods}
\label{resluts}
We compare our method with previous state-of-the-art methods for image deblurring and conduct quantitative and qualitative comparisons on the benchmark datasets. Unless otherwise stated, the high-level vision model used here is a multi-label classification model \cite{ridnik2021asymmetric}. Firstly, we compare the performance with and without the proposed semantic prior based on the simplest UNet structure. In addition, we embed the proposed prior into other recent mainstream image deblurring frameworks which have different structures, such as DeepDeblur \cite{nah2017deep}, MPRNet \cite{Zamir2021MPRNet}, MIMO-UNet \cite{cho2021rethinking}, and MSSNet \cite{Kim2022MSSNet} to verify the effectiveness of our proposed semantic priors. \textit{We use the same training strategy as those methods to train or fine-tune the models except for the total training iteration, which is time-consuming.} Meanwhile, the performances of several classical deblurring methods \cite{tao2018scale-recurrent,zhang2019deep,suin2020spatially,gao2019dynamic,NIU202169} are also listed. Public implementations with default parameters were used to obtain the results on test images. Some methods' codes are unavailable, then we use the results reported in their papers.

\begin{table}[!t]\small
\centering
\caption{Quantitative evaluation on the GoPro test dataset \cite{nah2017deep}. SP means the proposed semantic prior. The models with $\dagger$ indicate the simple version of the original methods. $\ddagger$ represents fine-tuning the models from the pre-trained models. Bold font indicates the results of using the proposed semantic priors.}
\label{tab:GoPro}
\begin{tabular}{l|c|c|c|c|c}
\toprule
Methods     & Structure & PSNR↑   & SSIM↑  & LPIPS↓        &  Time (s)                                  \\
\midrule
SRN \cite{tao2018scale-recurrent} &multi-scale & 30.26 & 0.934 & 0.136    & 0.736                            \\
PSS-NSC \cite{gao2019dynamic} &multi-scale   & 30.92 & 0.942 & 0.122               & 0.316            \\
DMPHN \cite{zhang2019deep} &multi-patch    & 31.20 & 0.945 & 0.128            & 0.307                    \\
SAPHN \cite{suin2020spatially} &multi-patch     & 31.85 & 0.948 & 0.101       & 0.340                        \\
Niu et.al. \cite{NIU202169} &-   &31.25 &0.945 &- &-                                         \\
\hline
\hline
UNet \cite{depthLi}    &\multirow{2}{*}{multi-scale}   & 30.29      & 0.939      &0.085  &0.086 \\                    
\textbf{UNet+SP}     &  &\textbf{31.06}       & \textbf{0.946}      &\textbf{0.075}   &\textbf{0.170} \\    
\hline 
$^\dagger$DeepDeblur \cite{nah2017deep} &\multirow{2}{*}{\makecell[c]{multi-scale}}  & {27.28}      & {0.832}      &  {0.247}     &{0.117}                          \\
\textbf{$^\dagger$DeepDeblur+SP} &  & \textbf{27.45}      & \textbf{0.835}      &  \textbf{0.237}     &\textbf{0.199}                          \\ \hline
$^\dagger$MPRNet \cite{Zamir2021MPRNet}  &\multirow{2}{*}{multi-patch}    & 29.90 & 0.936  & 0.135  &0.031\\
\textbf{$^\dagger$MPRNet+SP}  &  & \textbf{30.88}      & \textbf{0.946}      & \textbf{0.122}   &\textbf{0.105}\\
\hline
MIMO-UNet \cite{cho2021rethinking}  &\multirow{4}{*}{multi-scale}  &30.91   & 0.946  &0.119          & 0.019                \\
\textbf{MIMO-UNet+SP} &  & \textbf{31.08}       & \textbf{0.948}       & \textbf{0.113}       & \textbf{0.043}                         \\
MIMOUNetPlus \cite{cho2021rethinking} &  & 32.45 & 0.957 & 0.092        &0.309                 \\
\textbf{$^\ddagger$MIMOUNetPlus+SP} &  & \textbf{32.55}      & \textbf{0.962}      &  \textbf{0.089}     &\textbf{0.314}                          \\ 
\hline
MSSNet \cite{Kim2022MSSNet}  &\multirow{2}{*}{multi-stage} & {32.98}      & {0.965}      &  {0.087}     &{0.166}                          \\
\textbf{MSSNet+SP}  &   & \textbf{33.22}      & \textbf{0.967}      &  \textbf{0.084}     &\textbf{0.335}                          \\ 
\bottomrule
\end{tabular}
\end{table}

\subsubsection{{Quantitative results}}
\textbf{Results of GoPro.}
Table \ref{tab:GoPro} shows the performance of existing methods and without/with embedding the proposed semantic prior methods. Our baseline model UNet \cite{depthLi} achieves 30.29dB in PSNR, 0.939 in SSIM, and 0.085 in LPIPS, which is close to the early classic methods \cite{tao2018scale-recurrent}. \textit{UNet+SP} represents embedding the prior into the UNet. By doing so, the performance gets a considerable improvement in every restoration quality metric, especially +0.77dB in PSNR. With the proposed priors, we train a simple $^\dagger$MPRNet \cite{Zamir2021MPRNet} with the \textit{stage1} model and $^\dagger$DeepDeblur \cite{nah2017deep} with 6 ResBlocks, and retrain the MIMO-UNet \cite{cho2021rethinking} and MSSNet \cite{Kim2022MSSNet}, and finetune the $^\ddagger$MIMOUNetPlus \cite{cho2021rethinking} models to verify the generalization ability. A similar phenomenon can also be found in plugging the prior into these frameworks. Guided by our comprehensive semantic priors, image deblurring performance can be significantly improved (+0.98dB in MPRNet, +0.17dB in MIMOUNet, and +0.24 in MSSNet) compared to the models without using the priors. Runtime is measured by using the released code of each method to run the entire GoPro test dataset \cite{nah2017deep}. Then, the average running time (second) with a single Nvidia 3090 GPU is obtained. The running time after adding our priors is still within the reasonable range. The increase in running time is due to the operation of the $F_S$ when embedding the semantic priors.

From the results, the proposed semantic prior performs well in different deblurring structures, such as multi-scale, multi-patch, and multi-stage frameworks. Considering the U-shape model becomes the main-stream framework for image deblurring, we can evaluate our method with an early model DeepDeblur \cite{nah2017deep} to demonstrate the effectiveness on non-UNet models, which is stacked by several ResBlocks. The proposed priors also bring a significant  improvement compared to the original model.
\begin{table}[!t]\small
\centering
\caption{Image deblurring results on HIDE \cite{Shen2019HumanAwareMD} and RealBlur\_J \cite{rim2020real} dataset. The results of the HIDE are evaluated by using the models trained on GoPro \cite{nah2017deep}. The RealBlur\_J's results are trained and tested on the RealBlur dataset.}
\label{tab:HIDE}
\begin{tabular}{l|c|c|c|c}
\toprule
\multirow{2}{*}{Methods} &\multicolumn{2}{c|}{HIDE} &\multicolumn{2}{c}{RealBlur\_J} \\ \cline{2-5}
              & {PSNR↑} & {SSIM↑}  & {PSNR↑} & {SSIM↑}  \\
\midrule
SRN \cite{tao2018scale-recurrent}  & 28.36 & 0.915   & 31.38     & 0.909                              \\
DMPHN \cite{zhang2019deep}     & 29.09  & 0.924      &-- &--  \\
DeblurGAN-v2 \cite{kupyn2019deblurgan}  &-- &-- &29.69 &0.870 \\
\hline
UNet \cite{depthLi}        & 27.92     &0.915  & 28.98 &0.924          \\
\textbf{UNet+SP}       & \textbf{28.51}      & \textbf{0.924}    & \textbf{29.53} &\textbf{0.932}         \\
$^\dagger$MPRNet \cite{Zamir2021MPRNet}     & 28.41 & 0.922 & 29.41       &0.933       \\
\textbf{$^\dagger$MPRNet+SP} & \textbf{28.89}      & \textbf{0.930}  &  \textbf{29.54}  &\textbf{0.935}          \\
MIMOUNet+\cite{cho2021rethinking} & 29.99 & 0.930  & 29.44  & 0.932   \\
\textbf{$^\ddagger$MIMOUNetPlus+SP} & \textbf{30.05}      & \textbf{0.945}  & \textbf{29.67}      &\textbf{0.934}       \\
\bottomrule
\end{tabular}
\end{table}

\textbf{Results of HIDE and RealBlur.}
To verify the generalization of our methods to other scenarios, we directly apply the models trained on the GoPro to the HIDE dataset. The results are shown in Table \ref{tab:HIDE}. With our semantic priors, the PSNR and SSIM are improved effectively, which illustrates the robustness of our methods.
Table \ref{tab:HIDE} also shows the quantitative comparison results on the RealBlur dataset. Our training iteration is much less than the normal training strategy, thus, the final metrics in terms of PSNR are lower than SRN \cite{tao2018scale-recurrent} and DeblurGAN-v2 \cite{kupyn2019deblurgan}. Even in such real-world blurry scenarios, the results using our semantic priors (+SP) are still better than those without the prior with a steady improvement. 

\begin{table}[tb]\small
\centering
\caption{Improvements of recent SOTA methods on PSNR (dB), including results of baseline and baseline with their proposed modules. The last three rows represent the results with our proposed semantic priors on the GoPro dataset. Impro. indicates the improvements compared to the baseline model.}
\label{tab:Improvements}
\begin{tabular}{l|c|l|c}
\toprule
{Models}     &Baseline           &Baseline+proposed modules &Impro.\\
\midrule
{HINet-REDS\cite{chen2021hinet}} & 28.11   & +HIN: 28.23 &+0.12 \\
{HINet-GoPro\cite{chen2021hinet}} & 30.98  & +HIN: 31.40 &+0.42 \\
MPRNet\cite{Zamir2021MPRNet} &29.86  & +SAM+CSFF: 30.49 &+0.63 \\
MIMOUNet\cite{cho2021rethinking} &31.16  & +MIMO+AFF+MSFR: 31.73 &+0.57\\
\hline
Ours-UNet    & 30.29 &+ SP: 31.06 &+0.77 \\
Ours-MPRNet    & 29.90 &+ SP: 30.88 &\textbf{+0.98} \\
Ours-MIMOUNet    & 30.91 &+ SP: 31.08 &+0.17 \\
Ours-MSSNet    & 32.98 &+ SP: 33.22 &+0.24 \\
\bottomrule
\end{tabular}
\end{table}

\begin{figure*}[!t]
    \centering
    \includegraphics[width=4.6in]{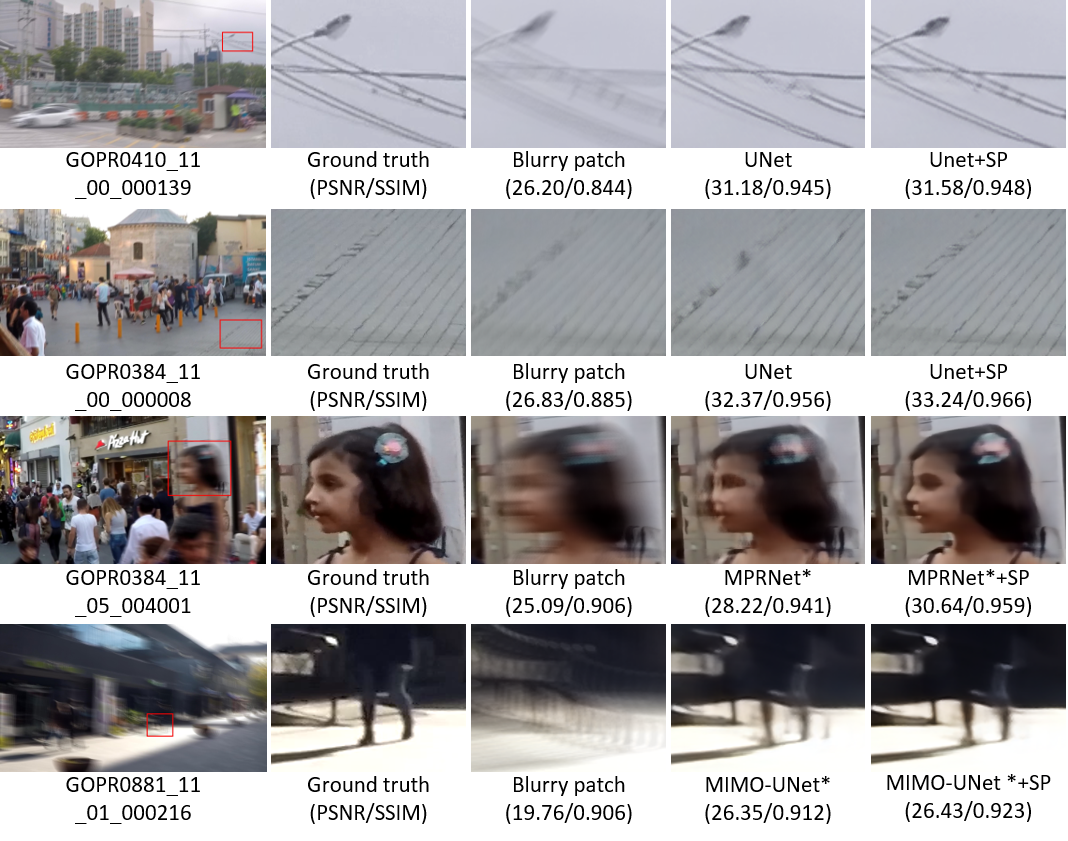}
    \caption{Visual comparisons of the deblurred images on GoPro \cite{nah2017deep} without and with proposed semantic prior (SP). We report the PSNR/SSIM in parentheses. Severe blur can be reduced with the guidance of proposed priors.}
    \label{fig:gopro}
\end{figure*}

\textbf{Improvements compared to the previous methods.}
Our results with the proposed semantic priors are all executed on the SOTA models, such as MPRNet, MIMOUNet, and MSSNet, which already have got high performance. However, we obtained good gains on several datasets compared to these models. Our most significant improvement on the GoPro dataset is +0.98dB. We collected the performance (PSNR) of the recent SOTA methods on image deblurring, including results of baseline and baseline with their proposed modules, which are shown in Table \ref{tab:Improvements}. From the tables, even the SOTA methods (MPRNet) only achieve a maximum gain of 0.63dB, which is lower than our improvements (0.98dB). Our method improves a lot more than previous SOTA methods. Therefore, we believe these improvements are relatively sufficient to validate our method in the image deblurring field. In addition, we also provide the results of SSIM and LPIPS in Table \ref{tab:GoPro}. The LPIPS of the MPRNet is improved by 10.66\% (from 0.135 to 0.122, lower is better) with semantic priors on the GoPro dataset (shown in Table \ref{tab:GoPro}), indicating the proposed method's effectiveness in the visual perception view.
\begin{figure}[!t]
    \centering
    \includegraphics[width=4.6in]{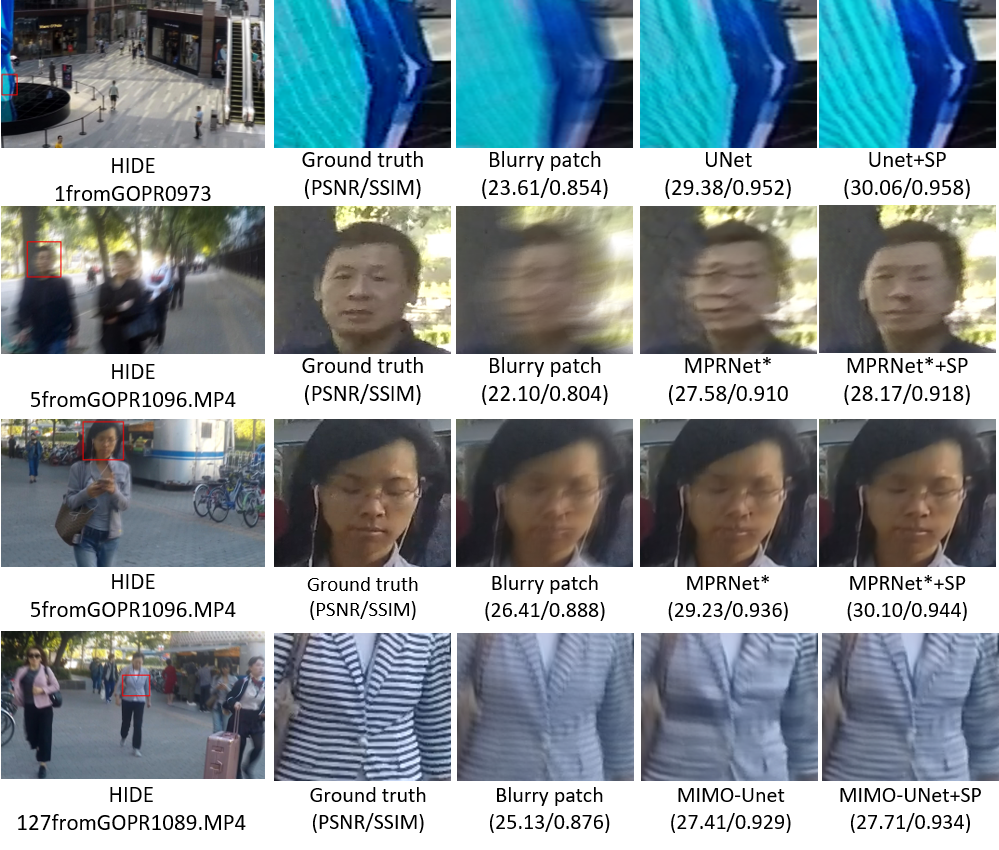}
    \caption{Visual comparisons on HIDE \cite{Shen2019HumanAwareMD}. The evaluation models are only trained on the GoPro dataset \cite{nah2017deep}. 
    }
    \label{fig:hide}
\end{figure}

\subsubsection{\textbf{Qualitative results}}
Figure \ref{fig:gopro} and \ref{fig:hide} show some qualitative results of deblurring methods on GoPro \cite{nah2017deep} and HIDE \cite{Shen2019HumanAwareMD}.  
After embedding our priors into different deblurring frameworks, such UNet, MPRNet, and MIMO-UNet, the deblurred images obtain clear contents, and the severe blur is reduced meaningfully, see the last column of figures. The deblurred images with the proposed priors, which contain more detailed structures, are clearer than the baseline results. The performance below the pictures also demonstrates superiority.
\begin{figure*}
    \centering
    \includegraphics[width=5.3in]{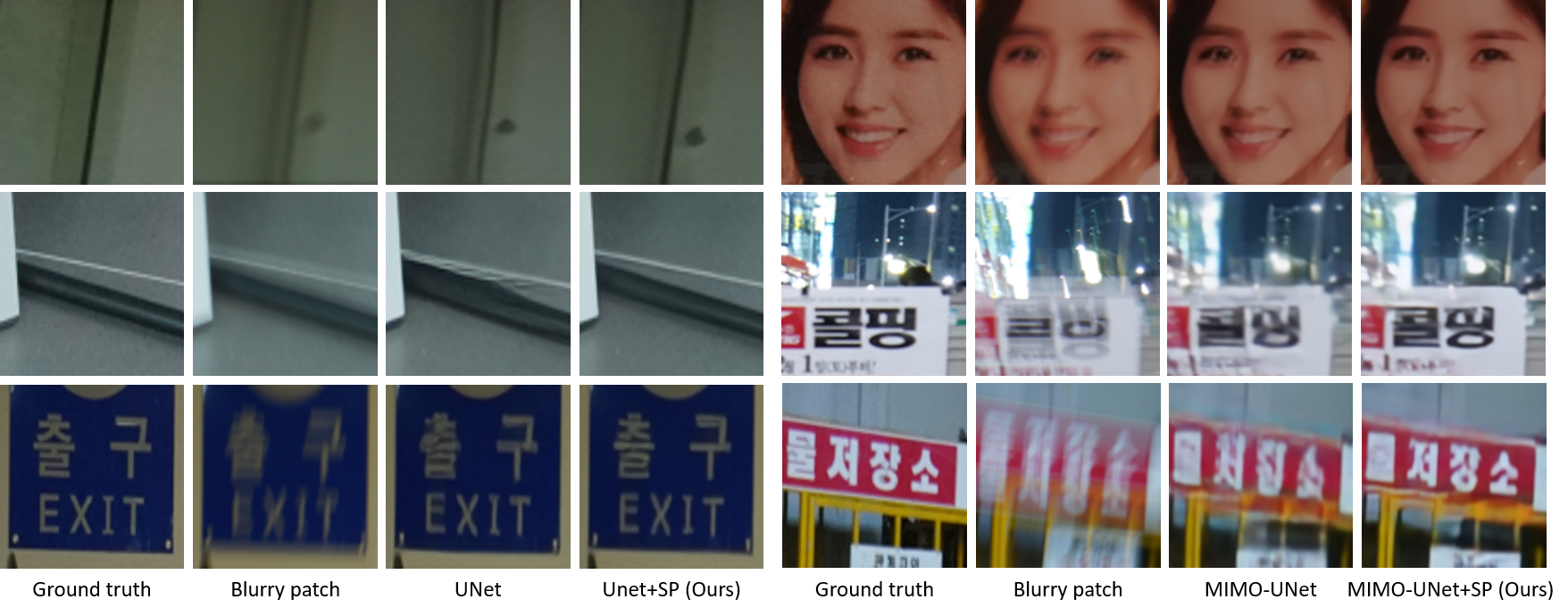}
    \caption{Deblurred results without and with the semantic prior on the RealBlur \cite{rim2020real} dataset. +SP represents the deblurred results with our learned semantic priors. Using our priors reduced the severe blur and deblurred images retain fewer artifacts.}
    \label{fig:realblur}
\end{figure*}
Most recent works perform well on the synthetic dataset but may fail in real-world severe blur images while handling severe blur is a considerable challenge. However, our method focuses on severe blur removal, which significantly improves visuals.  Fig. \ref{fig:realblur} gives some results of real blurry images. Whether the baseline model is UNet or MIMO-UNet \cite{cho2021rethinking}, embedding semantic priors outperform without priors in several cases, including text and face images, even in the low-light or extremely light conditions.

\section{Ablation studies and Discussions}
\label{sec:ablation}
In this section, we further verify the key designs of our semantic prior learning branch and prior embedding layer. Table \ref{tab:spl} shows the average performance of ten times evaluations of different partitions of the SPL branch.

\begin{table}[tb]\small
\centering
\caption{Ablation study of our model on the GoPro test dataset \cite{nah2017deep}. Net0 represents the baseline UNet model (only IDe branch). Net0* refers to the model being trained with SPL but didn't use distillation loss. -- indicates that the training is unstable.}
\label{tab:spl}
\begin{tabular}{c|c|c|c|c|c|c}
\toprule
Models &HCL & CLC & MLA & SAT  & PSNR↑ &LPIPS↓     \\ 
\midrule
Net0    &\xmark  & \xmark  & \xmark   & \xmark                            & 30.2886 &0.0850 \\ 
Net0*    &\xmark      & \cmark & \cmark   & \cmark                         & -- & -- \\ 
\hline
Net1    &\cmark     & \xmark    & \xmark  & Add                        & 30.6501 &0.0812 \\
Net2    &\cmark     & \xmark    & \xmark  & Concat                        & 30.7331 &0.0790 \\
Net3    &\cmark     & \xmark    & \xmark   & \cmark                      & 30.7712 &0.0785 \\
Net4    &\cmark     & \cmark  & \xmark      & \cmark                  & {30.9648} &{0.0776}   \\
Net5    &\cmark     & \xmark     & \cmark   & \cmark                  & 30.8705 
& 0.0771 \\
Net6    &\cmark     & \cmark    & \cmark   & \cmark                   & \textbf{31.0608} & \textbf{0.0745}  \\ 

\bottomrule
\end{tabular}
\end{table}

\begin{figure*}[htb]
    \centering
    \includegraphics[width=5.4in]{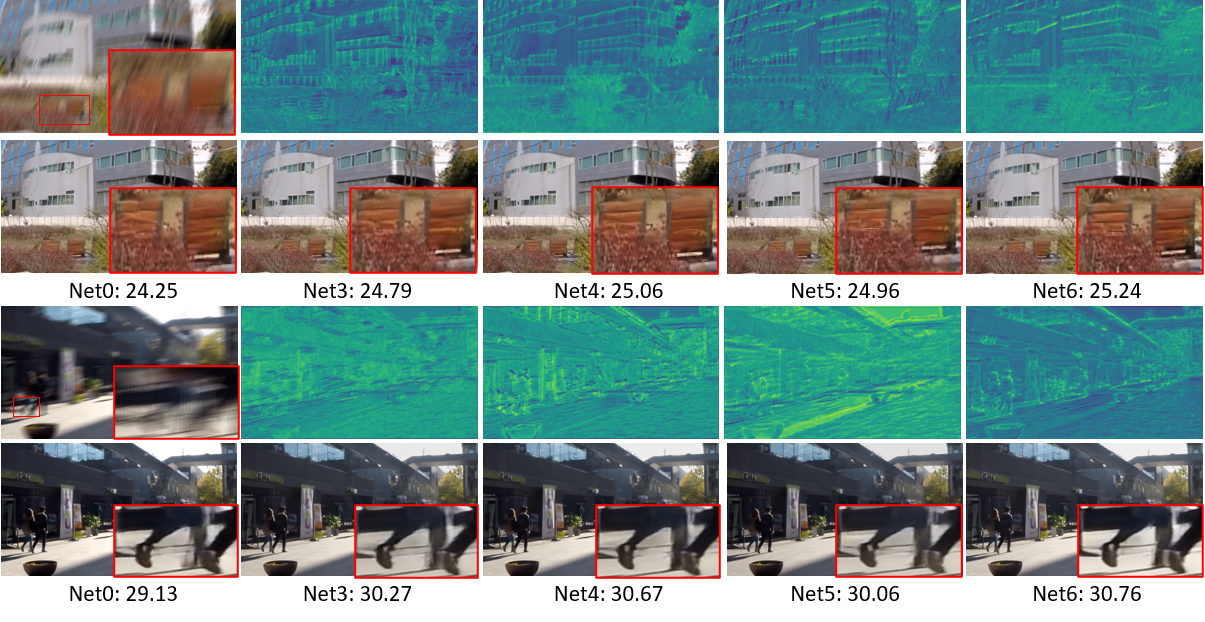}
    \caption{Deblurred results (bottom row) and the learned middle-level priors (top row) with different versions of the proposed method. The priors of the entire image are shown with color maps for a clear view of the semantics. Numbers indicate the PSNR value.}
    \label{fig:visual_spl}
\end{figure*}

Net0 represents the baseline UNet model without using any semantic prior or any parts of the SPL branch. Net0* means that training with the $F_S$ but without $L_{prior}$ supervision. However, training without $L_{prior}$ leads to unstable results. After adding feature distillation HCL loss (Eq. (\ref{eq:fea})), the performance gains a lot, which can be seen from Net0 and Net1. Except for using the proposed SAT module to embed the priors to the IDe branch (Net3), using simple \textit{addition} and \textit{concatenation} can also improve the performance compared to Net0, which indicates the effectiveness of the proposed semantic prior in the image deblurring task. With the primary feature distillation, we can obtain an initial semantic prior with crucial sharp structures and a few abstract contexts to improve the baseline performance, which demonstrates the effectiveness of the feature distillation.  After using CLC to enhance the semantics of the generated priors, the deblurring performance achieves a higher result (Net4), which indicated that richer global context information can guide the discrimination of the objects in severe blur scenarios. Introducing the MLA from different scales leads Net5 to perform better than Net3, indicating the importance of combining the sharp local structure and the global semantic information. Net6 consists of all modules we proposed, the learned semantic prior yields better capabilities in PSNR and LPIPS, which demonstrates the superiority of our method in pixel-wise measurements and the human perception field. 

Fig. \ref{fig:visual_spl} shows the visualized priors learned by the different parts of the SPL. First, the deblurred images with any priors learned by SPL outperform those without semantic priors (Net0). Additionally, as we adopt the feature distillation loss, the severe blur is removed meaningfully, and the correct structures of deblurred images are recovered progressively. The priors in each top row demonstrate the proposed HCL loss enabling the priors to learn the sharper local textures (Net3). CLC leads the shallow-level priors to acquire deep-level information and enhance the global contexts with object-level semantics (Net4). The priors from the Net5 show the aggregated priors by MLA, resulting in a finer deblurred image. With all parts of our methods, the deblurred image has more clear contents and structures, and the priors discriminate the contexts obviously (Net6). The priors of Net4 and Net6 show significant semantics, such as grass, building, and sky, indicating the superior ability of CLC to capture semantics.

\textbf{Learning priors with different high-level vision tasks.} We explore the effects of the different high-level vision tasks and the prior types. We adopt the classification and segmentation models to see the deblurring results of different tasks, which are the main tasks in the high-level vision area. We investigate the semantic prior for the segmentation task both at the \textit{image-level} and \textit{feature-level}. To apply \textit{image-level prior}, similar to \cite{Shen2018DeepSF}, we train an individual segmentation model with the blurry input, which is supervised by a pre-trained segmentation model \cite{zhu2019improving} under the knowledge distillation strategy. By doing so, the useful information from the sharp images can be transferred to generate the segmentation map (probability or one-hot map), which can be regarded as the image-level semantic prior. Feature-level prior is the multi-level semantic priors, as introduced in Sec. \ref{sec:approach}.

 \begin{figure}[!t]
    \centering
    \includegraphics[width=4.0in]{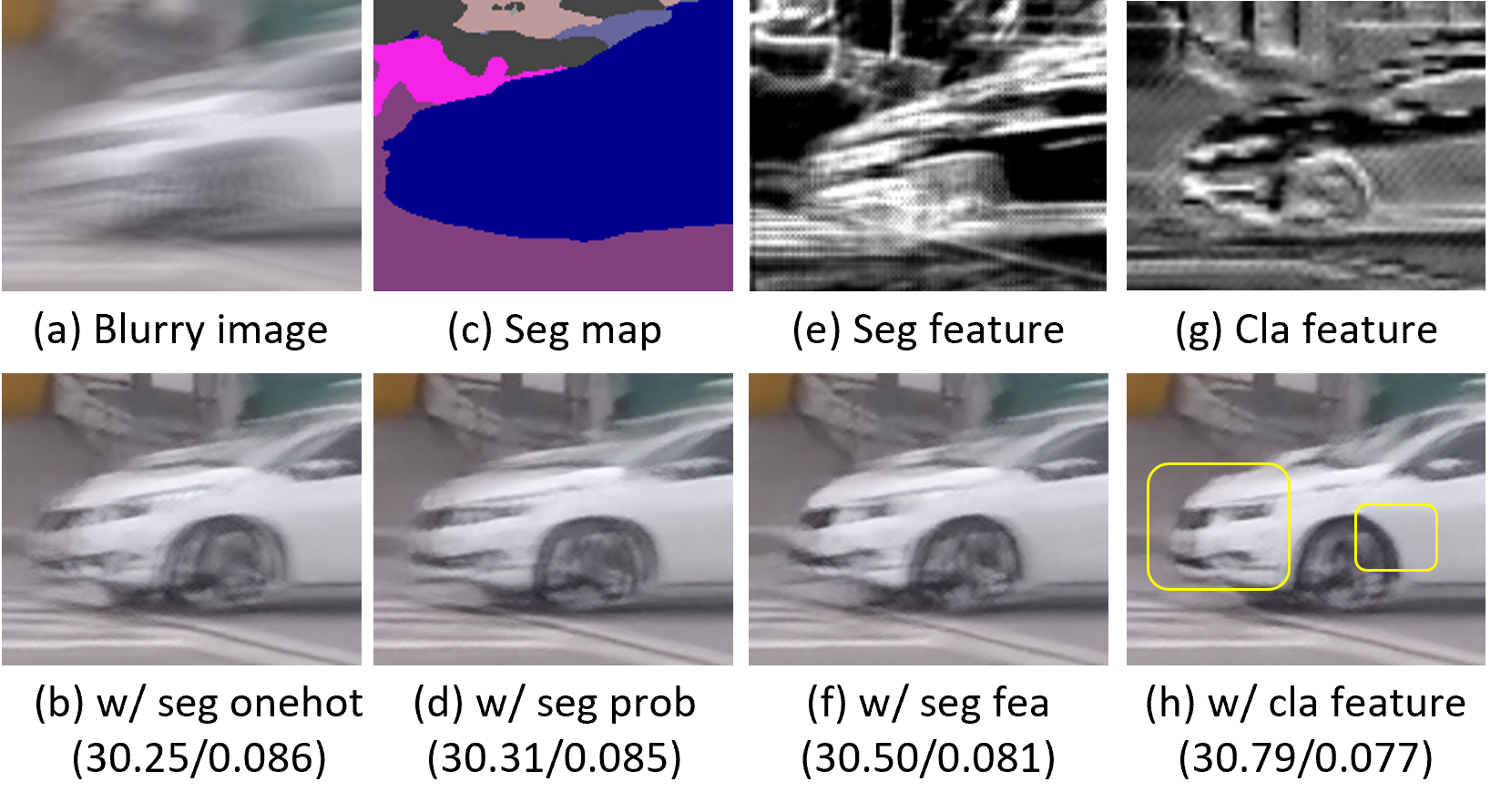}
    \caption{Deblurred results using segmentation (seg) and classification (cla) models with image and feature-level priors. Blue and purple regions in (c) indicate the segmentation category of car and ground according to \cite{cordts2016cityscapes}. We report the average PSNR/LPIPS in parentheses. Zoom in for detail.}
    \label{fig:seg_prob}
\end{figure}
We show the deblurred results with the different prior types in Fig. \ref{fig:seg_prob}. The features are averaged by all channels of priors.
Due to the sparse information of the segmentation map (c), the deblurred results rely on the one-hot (b) or the probability map (d) is rather disappointing. Although the probability map contains more details than one-hot, its results still perform worse than the feature-level prior. On the contrary, \textit{feature-level} prior can provide additional structure information, see Fig. \ref{fig:seg_prob} (e) and (g). Combining global and local knowledge, the deblurred results of (f) and (h) are better than (b) and (d) which are generated under the \textit{image-level} prior guidance. We also find that with the guidance of the classification task (g), the learned priors have more clear structure than segmentation (e), which leads to a more accurate result, close to the sharp image. The performance (PSNR/LPIPS) below the images also demonstrated that the classification model is more suitable for our deblurring problem. An intuitive explanation is the segmentation focuses more on the shape and edge instead of the object's context, which is insufficient for supporting severe blur detection and removal. 
\begin{figure}[!t]
    \centering
    \includegraphics[width=4.0in]{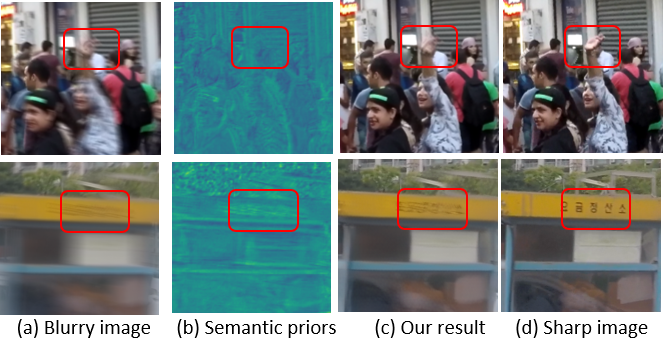}
    \caption{Failure case due to the out-plane hands shaking (top row) and tiny structures in a specific semantic region (bottom row).}
    \label{fig:limitate}
\end{figure}
\textbf{Limitations and future works.}
The proposed semantic prior guided image deblurring method can handle the severe blur removal problem effectively, which introduces a complementary clue to assist in distinguishing the objects in the severely blurry regions. However, due to the depth change, our method can not obtain a good result for some out-plane motion blur, such as the shaking hands in the top row of Fig. \ref{fig:limitate}. Although the hand can be classified by the proposed priors, this depth-changing blur is still hard to remove. In addition, when there are some very tiny structures in a specific region, such as the bottom patch in Fig. \ref{fig:limitate}. Because the characters are too small and the degree of blur is very large, and the category of the classification task cannot cover these tiny structures, these very detailed texts are hard to learn semantics. There still has a performance gap between our method and the ground truth image. Therefore, we will focus on 1) out-plane movement caused blur, and 2) explore the new technique to handle the tiny objects deblurring using semantic information in the future. 

\section{Conclusions}
\label{sec:conclusion}
Previous deblurring works may fail when meeting the severely blurry regions. This paper focuses on severe blur removal in general scenes by learning a comprehensive semantic prior from the high-level vision tasks. The sharp semantic prior, including global context and local clear structures, can be learned through a cross-level feature distillation loss. A semantic prior embedding layer plugs an aggregated prior into the image deblurring branch to realize the blur removal. The experiments and ablation studies prove the effectiveness of our proposed method. 


\section*{Acknowledgments}
This work is supported by the Project of the National Natural Science Foundation of China (Grant No.61901384 and No.61871328), as well as the Joint Funds of the National Natural Science Foundation of China (Grant No.U19B2037).

 \bibliographystyle{elsarticle-num} 
 \bibliography{mybib.bib}





\end{document}